\documentclass[10pt,twocolumn,letterpaper]{article}

\usepackage[pagenumbers]{cvpr} %
\usepackage{float}
\usepackage{algorithm}
\usepackage{algorithmic}
\usepackage{amsmath}
\usepackage{graphicx}
\usepackage[symbol]{footmisc}

\definecolor{cvprblue}{rgb}{0.21,0.49,0.74}
\usepackage[pagebackref,breaklinks,colorlinks,citecolor=cvprblue]{hyperref}

\def\paperID{16384} %
\def\confName{CVPR}
\def\confYear{2025}

\begin{document}

\title{Real-time One-Step Diffusion-based Expressive Portrait Videos Generation}

\author{
Hanzhong Guo$^1$\footnotemark[2]\,\ \footnotemark[1] \quad Hongwei Yi$^2$\footnotemark[2],\footnotemark[3]  \quad Daquan Zhou$^{3}$  \\
Alexander William Bergman$^2$\quad  Michael Lingelbach$^{2}$\quad  Yizhou Yu$^{1}$\footnotemark[4] \\ 
{\normalsize 
\quad $^1$HKU
\quad $^2$Hedra Inc \quad $^3$NUS}
}
\vspace{-1em}
\twocolumn[{%
\renewcommand\twocolumn[1][]{#1}%
\maketitle
\includegraphics[width=.95\linewidth]{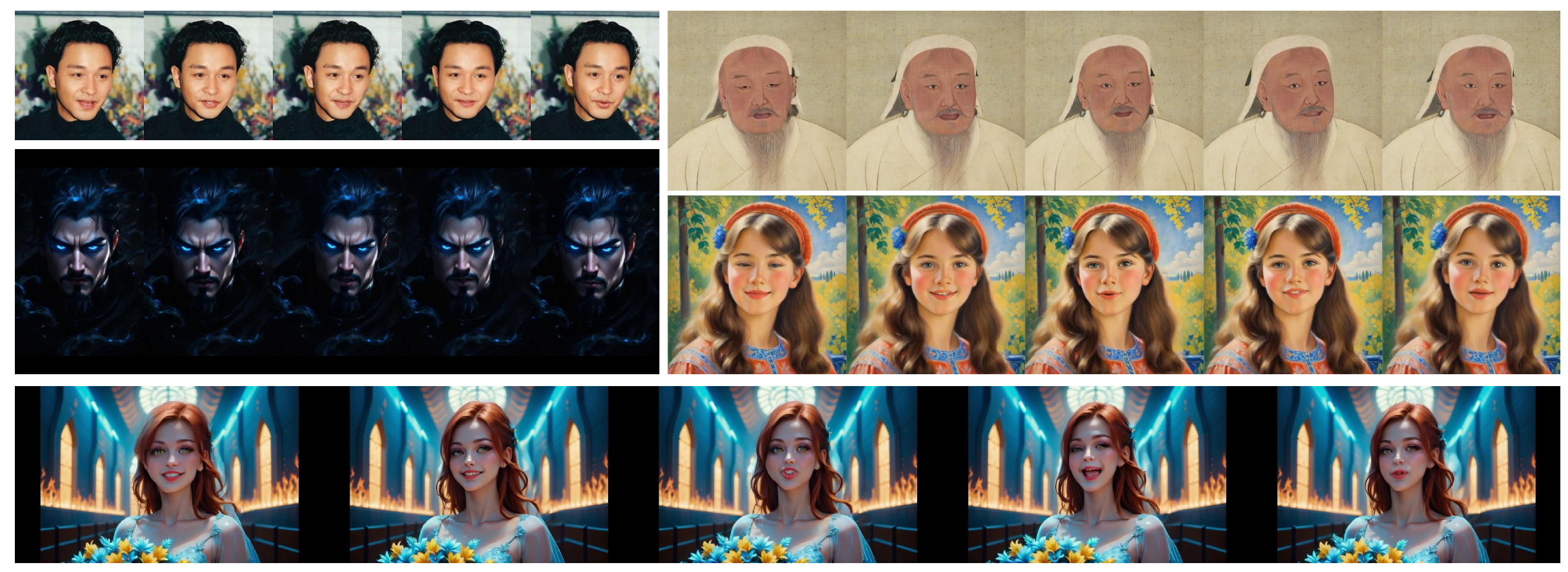}
\captionof{figure}{One-step generation results for our proposed one-step avatar latent consistency models (OSA-LCM). We can generate high-quality and diverse resolutions (512x512, 512x896, 896x512) with a single sampling step.  \vspace{1em}}
\label{fig:teaser}
}]
\maketitle
\footnotetext[1]{Work done during an internship at Hedra Inc.}
\footnotetext[2]{Equal contribution, $^\ddagger$Project lead, $^\S$Corresponding author.}
\begin{abstract}
Latent diffusion models have made great strides in generating expressive portrait videos with accurate lip-sync and natural motion from a single reference image and audio input. However, these models are far from real-time, often requiring many sampling steps that take minutes to generate even one second of video—significantly limiting practical use.
We introduce OSA-LCM (One-Step Avatar Latent Consistency Model), paving the way for real-time diffusion-based avatars.
Our method achieves comparable video quality to existing methods but requires only one sampling step, making it more than 10x faster. 
To accomplish this, we propose a novel avatar discriminator design that guides lip-audio consistency and motion expressiveness to enhance video quality in limited sampling steps. 
Additionally, we employ a second-stage training architecture using an editing fine-tuned method (EFT), transforming video generation into an editing task during training to effectively address the temporal gap challenge in single-step generation.
Experiments demonstrate that OSA-LCM outperforms existing open-source portrait video generation models while operating more efficiently with a single sampling step.
More results are available at \url{https://guohanzhong.github.io/osalcm/}.
\end{abstract}
    
\section{Introduction}
\label{sec:intro}

Large-scale diffusion models have made great process in generating high-fidelity images and video~\cite{ramesh2021zero,rombach2022high,saharia2022photorealistic,podell2023sdxl,esser2024scaling,brooks2024video}, often with controllable ability.
The current controllability of large-scale diffusion models allows single image input to control image and video generation~\cite{zhang2023adding}, and several reference images input to achieve customized image and video generation~\cite{ruiz2023dreambooth,jiang2024videobooth}. In addition, some studies have investigated video-driven or audio-driven video generation, e.g., input audio-driven head animation, or portrait video generation~\cite{sun2023vividtalk,tian2024emo,chen2024echomimic,wei2024aniportrait,xu2024hallo}.

Portrait video generation aims to generate a talking head video given input audio, requiring the video to have sufficient facial expressions and audio-aligned lip movements.
Some works use a 3D morphable model (3DMM) as the intermediate representation and use the audio to drive the 3DMM to generate portrait video~\cite{zhang2023sadtalker}.
For example, Vividtalk~\cite{sun2023vividtalk} uses blendshapes for the intermediate representation. 
However, these methods are all constrained by the intermediate representations. The videos generated by these methods lack overall expressiveness and naturalness because they are constrained to the space represented by the 3DMM.
To tackle such challenges, some works used large-scale diffusion models to accomplish the portrait video generation benefiting from the prior diffusion models~\cite{tian2024emo,chen2024echomimic,wei2024aniportrait,xu2024hallo}. 

Incorporating diffusion techniques and parametric
or implicit representations of facial dynamics in a latent
space achieves the end-to-end generation of high-quality,
realistic animations.
These methods take the reference image and audio as conditions and train the diffusion models to predict the ground truth video. 
However, since the discretization process of diffusion models, methods based them are very slow. Often, it takes more than a minute to generate one second of portrait video, which makes it impossible to use it in practice.

There are lots of works that focus on accelerating the speed of image generation in diffusion models, via training-free methods~\cite{song2020denoising,bao2022analytic,lu2022dpm} or training-based methods~\cite{meng2023distillation,song2023consistency,luo2023latent}. 
Referring to the works in acceleration of image generation, some approaches utilize consistency distillation in the latent space (LCM) and can achieve competitive results in four sampling steps in video generation~\cite{wang2023videolcm,mao2024osv}.
However, most existing methods for accelerating the inference speed of video generation don't consider the characteristics of video generation, leading to blurry videos at one or two sampling steps.
This situation will become more serious in portrait video generation because, compared with general video generation, the range of motion of the head in the portrait video generation scene is larger. Meanwhile, the speed of motion is faster, and the blur or artifact problem becomes more serious within limited sampling steps.

This paper introduces the one-step avatar latent consistency model (OSA-LCM). This novel training framework transforms a single portrait image into an expressive video within one diffusion sampling step.
We train OSA-LCM with two stages of latent consistency model training.
In both stages, we design a novel discriminator for portrait video generation and train the adversarial latent consistency model (Adv-LCM) with both consistency loss and adversarial loss. 
In the first stage, we train the Adv-LCM with a normal training scheme, after which the Adv-LCM can generate high-fidelity portrait videos in two sampling steps.
In the second stage, we continuously train Adv-LCM using the editing fine-tuned methods (EFT) which  addresses the problem of a temporal gap in one sampling step. Trained OSA-LCM can generate the high-fidelity and natural video in only one sampling step, making it possible for the OSA-LCM to generate one second of video in close to one second.
As a result, the OSA-LCM greatly improves the speed of diffusion models in portrait video generation. Under one-step diffusion sampling, it can realize one-second video generation in one second while maintaining a similar generation effect from both quantitative and qualitative perspectives.

\begin{figure*}
  \centering
    \centerline{\includegraphics[width=0.96\linewidth]{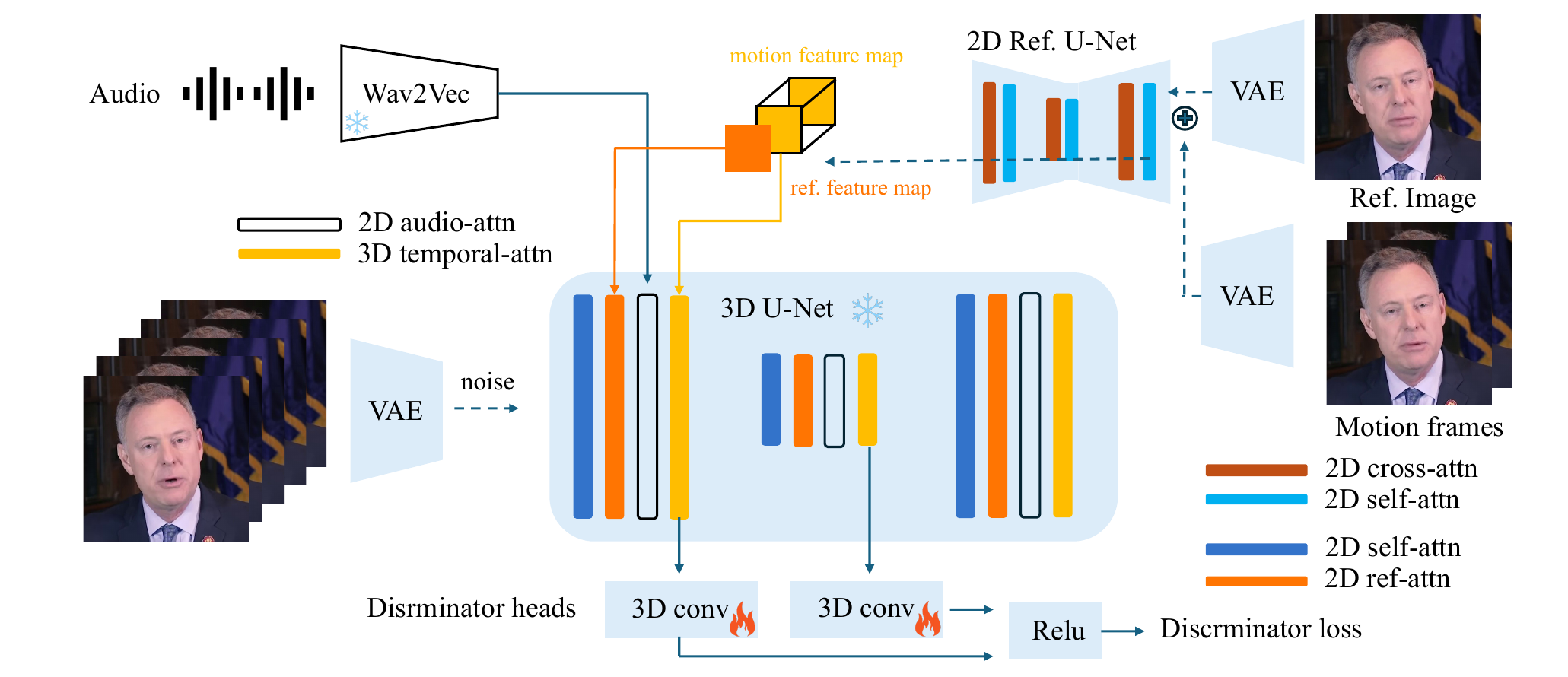}}

   \caption{\textbf{Model architecture for our base model and proposed discriminator.} The architecture of the base model refers to the EMO.}
   \label{fig:model}
   \vspace{-1em}
\end{figure*}

\section{Related works}
\subsection{Diffusion models}
Diffusion models~\cite{sohl2015deep,ho2020denoising,song2019generative,song2020score} consist of one forward process which noises the data into the Gaussian distribution gradually, defined as $x_t=q(x_0,\epsilon ,t) = \alpha_tx_0+\beta_t\epsilon, \epsilon  \sim \mathcal{N}(0,I)$. $\alpha_t$ and $\beta_t$ are the pre-defined noise scheduler. Diffusion models have one backward process which denoises the Gaussian distribution into the data via a learned neural network.
Recent works design numerical solvers of ordinary differential equations (ODE) or stochastic differential equations (SDE) to accelerate the diffusion sampling speed~\cite{bao2022analytic,song2020denoising,lu2022dpm}.

\subsection{Consistency models}
The consistency model (CM)~\cite{song2023consistency} is a novel family of generative models that enables few-step sampling. 
The core of CM is trained to take advantage of the fact that the points of the probability flow ODE (PF-ODE) trajectory converge to the same point, formulated as, for any timesteps $t,s \in [0,T], f(x_t,t)=f(x_s,s)$, where $f(\cdot)$ denotes the $x_0$ prediction function such as the PF-ODE sampling along the trajectory to the data and therefore CM has the boundary condition that $f(x_0,0)=x_0$.
Therefore, in order to meet the boundary condition, ~\citet{song2023consistency} parameterize the CM as, 
\begin{align}
\label{cm_eq} 
f_\theta(x_t,t) = c_{\text{skip}}(t) x_t + c_{\text{out}}(t) F_\theta(x_t,t),
\end{align}
where $F_\theta(\cdot)$ is the consistency models and $c_{\text{skip}}(\cdot)$, $c_{\text{out}}(\cdot)$ are differentiable functions~\cite{karras2022elucidating,song2023consistency}. It is noted that $c_{\text{skip}}(0)=1$, $c_{\text{out}}(0)=0$ to meet the boundary condition $f(x_0,0)=x_0$.

Meanwhile, CM has two training forms, consistency distillation (CD) and consistency training (CT), which have differences in whether to train the CM using the pre-trained diffusion models or not.
The CD has better performance than CT in a few-step sampling, which uses a pre-trained diffusion model called teacher model $\epsilon_\phi$ and a pre-defined ODE solver $\Phi$ to estimate the previous diffusion state in the empirical PF-ODE trajectory,
\begin{align}
\label{lcm_solver}
x_{t-\bigtriangleup  t}^\phi = x_t - \bigtriangleup  t \cdot  \Phi(x_t,t;\phi ),
\end{align}
and combined with the characteristic of the CM, we can train a student model $\theta$ which is $x_0$ prediction function,
\begin{align}
\label{loss_cd}
\mathcal{L}_{\text{CD} } = \mathbb{E}_{x_0,t,\epsilon } \left [ d(f_\theta(x_t,t),f_{\theta^-}(x_{t-\bigtriangleup  t},t-\bigtriangleup  t)) \right ]  ,
\end{align}
where $\theta^-$ is the EMA of function $f(\cdot)$, we take the distance function $d$ as the Huber loss~\cite{wang2024animatelcm} $d(x_t,x'_t) = \sqrt{\left \| x_t-x'_t \right \|_2^2 + \delta^2 }- \delta $, where $\delta$ is a threshold hyperparameter.
Meanwhile, since the large-scale diffusion models lie in diffusing in the latent space~\cite{rombach2022high}, latent consistency models (LCM)~\cite{luo2023latent} propose to train the consistency models in the latent space by replacing the pixel with the latent representation. 
For the simplicity of notion, all $x$ in this paper represent variables in the latent space.

\subsection{Portrait video generation}
The researches on portrait video generation focus on generating talking videos that match the audio input. 
There are two kinds of approaches to generating portrait videos, from end-to-end via a 2D neural network or from animating the 3D head models.
Sadtalker~\cite{zhang2023sadtalker} generates the corresponding 3D motion coefficients of the 3DMM from audio and then modulates a 3D-aware face render for achieving portrait video generation. Vividtalk~\cite{sun2023vividtalk} is modeled using blend shape and vertex as intermediate representations for coarse-grained and fine-grained, respectively. 
The generated 3D motion coefficients or intermediate representations are then mapped to the unsupervised 3D key points space of the proposed face referred to the audio and rendered to synthesize the portrait video~\cite{zhang2023metaportrait}.
However, these explicit methods are limited by the precision of 3D reconstructions and expressive capabilities.

Meanwhile, some works focus on generating portrait videos from end-to-end generative models. Wav2lip~\cite{prajwal2020lip} overlays synthesized lip movements onto existing video content while ensuring audio-lip synchronicity through the guidance of a discriminator. Recently, diffusion models have shown strong ability in vision generation. Some works~\cite{xu2024hallo,chen2024echomimic,wei2024aniportrait,tian2024emo,xu2024vasa} employed the pre-trained image or video diffusion models to generate portrait videos. They take the reference image and audio as two kinds of conditions and train extra modules to handle these conditions to predict the ground truth video, which can address
the challenges of lip synchronization, expression, and pose
alignment via one generative model. 
However, due to the discretization sampling characteristic of diffusion models, such methods take tens of times the length of the generated video to generate, making it impossible to use in practice.

\section{Model architecture}

Similar to the previous works~\cite{tian2024emo,xu2024hallo}, we train the base diffusion models starting from the Stable Diffusion v1.5 (SD v1.5) and add the temporal module in the Animatediff~\cite{guo2023animatediff}.

Since the current open-source portrait video diffusion models generate videos with smaller motions, we retrained a base model regarding the EMO~\cite{tian2024emo}.
Our proposed base model framework is depicted in Fig.~\ref{fig:model}, which consists of several modules.

\textbf{2D Reference U-Net.}
This module aims at preserving facial identity and background consistency. We utilize the backbone of SD v1.5. We concat the reference image and motion frames and encode them as the input of 2D Reference U-Net. %
For each layer of the output of the 2D Reference U-Net. we divide it into a reference feature map and a motion feature map and concatenate it with the input of the corresponding layer of the 3D Denoising U-Net.

\textbf{Audio Encoder.} 
We extract the audio embedding via the Wav2Vec model~\cite{schneider2019wav2vec} and then concatenate the audio embedding to the corresponding video. 
Similarly, we define the voice features of each generated frame by concatenating the features of nearby frames, $\mathcal{A}^f = \text{concat}(A^{f-m},...,A^{f},...,A^{f+m})$, where $A^{f}$ is audio embedding for the f-th frames, and m set as 2 in our paper.
We use an audio-atten layer in the 3D denoising U-Net which is a cross-attention block to effectively integrate audio features into the generation procedure.

\textbf{Temporal Attention Layer.}
Similar to the EMO, we apply self-attention temporal layers to the features within frames. Specifically, we reconfigure the input feature map with shape (b,c,f,h,w) to the shape (b*h*w,f,c) and conduct the self-attention to process the embedding.
Here, b is the batch size, h and w indicate the spatial dimensions of the feature map, f is the number of generated frames, and c is
the feature dimension. we merge the temporal layer inputs with the pre-extracted motion features of matching resolution along the frame dimension. For the generation of the first video clip, we initialize the motion frames as zero maps.
We adopt the same model architecture for the base model and OSA-LCM.

\section{OSA-LCM training method}

In this section, we will introduce the two training stages of OSA-LCM.
Both two stages train the adversarial latent consistency model (Adv-LCM) under the consistency loss and adversarial loss. 
We propose to train the Adv-LCM with different training paradigm definitions, which is like making the Adv-LCM progressively more capable of generating fewer steps using progressive training.

\subsection{Stage 1: Adversarial consistency distillation}
\begin{figure}[h!]
    \centering
    \begin{subfigure}[b]{0.15\textwidth}
        \centering
        \includegraphics[width=\textwidth]{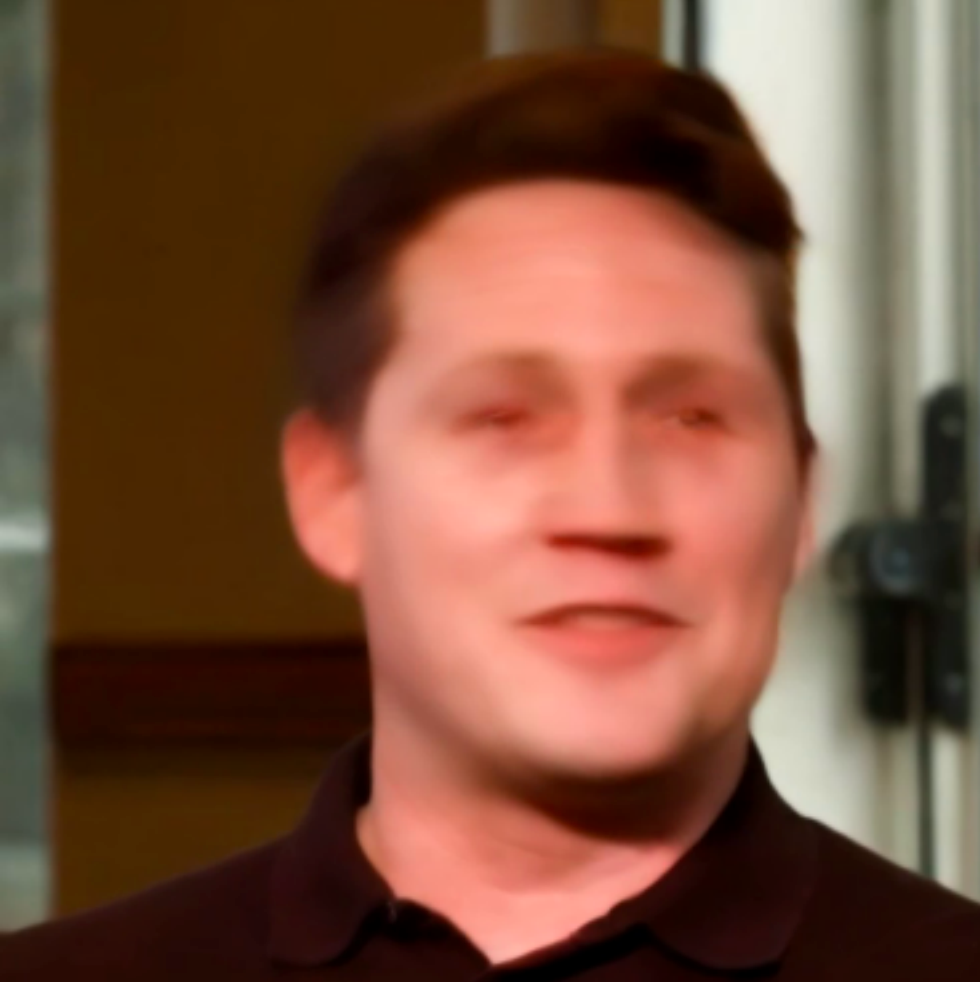}
        \caption{One step}
    \end{subfigure}
    \begin{subfigure}[b]{0.15\textwidth}
        \centering
        \includegraphics[width=\textwidth]{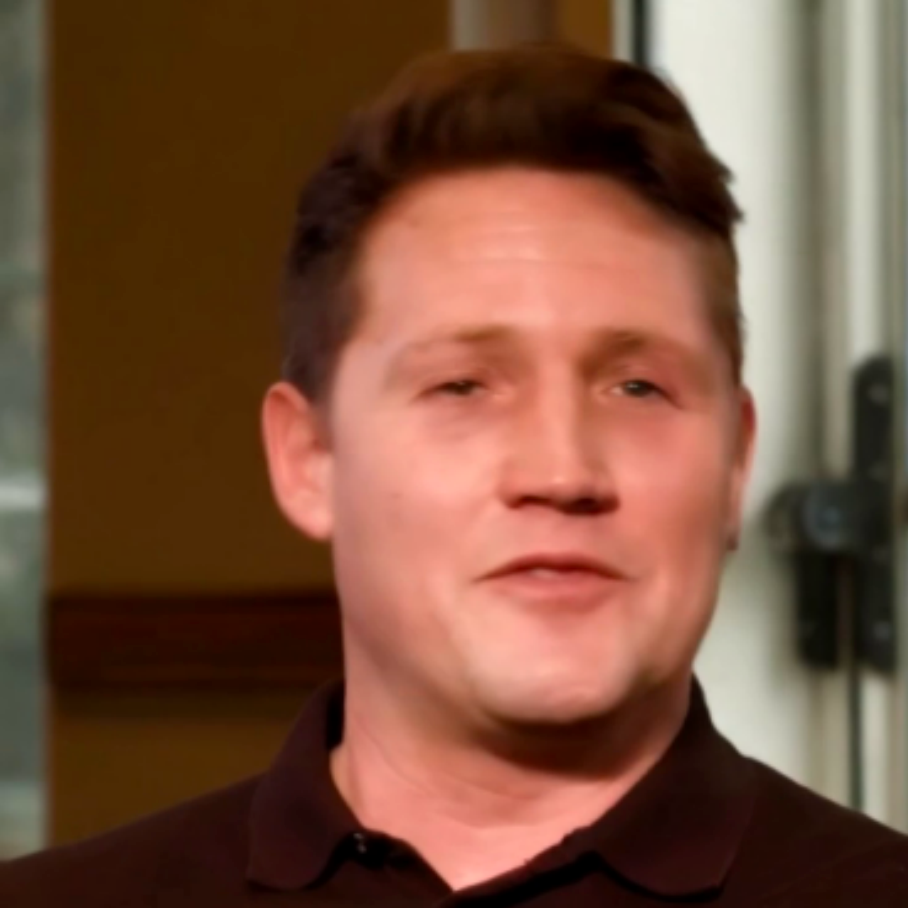}
        \caption{Two steps}
    \end{subfigure}
    \begin{subfigure}[b]{0.15\textwidth}
        \centering
        \includegraphics[width=\textwidth]{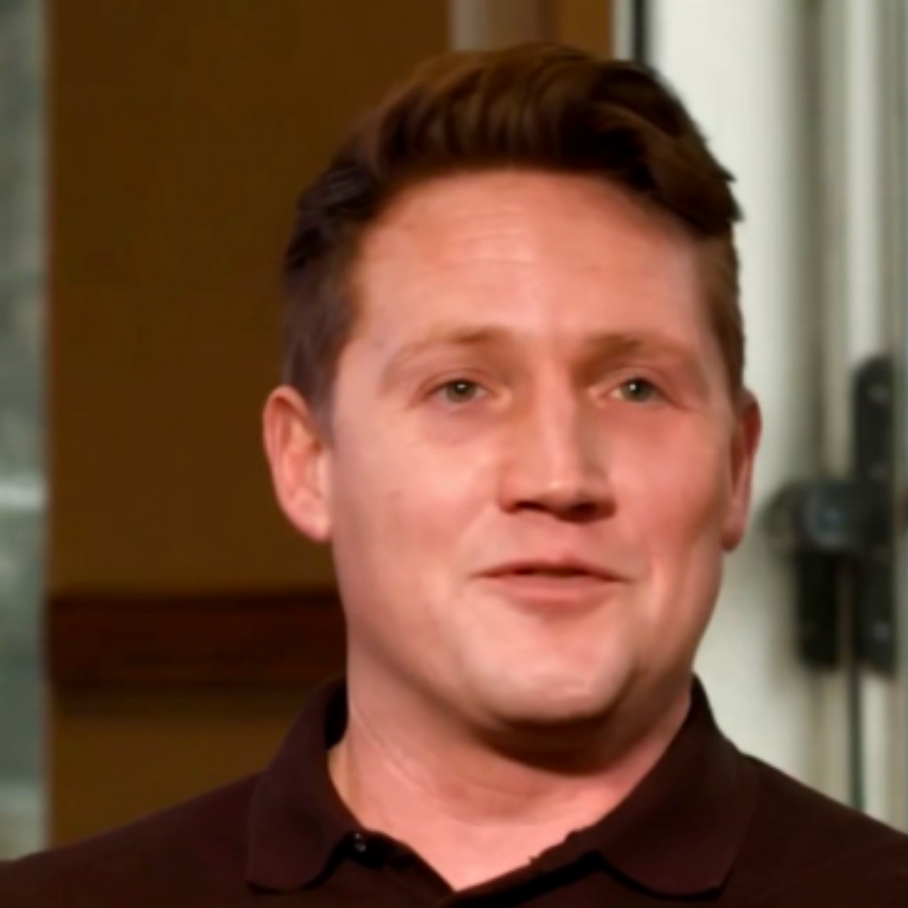}
        \caption{Four steps}
    \end{subfigure}
    \caption{Different sampling steps for vanilla LCM trained via only consistency distillation loss.}
    \label{fig:lcm}
\end{figure}

In the empirical study, we found LCM trained in Eq.~\eqref{loss_cd} can generate high-fidelity portrait videos with four or more sampling steps. However, when the sampling steps to two or one, the generated portrait videos become blurry and prone to artifacts, shown in Fig.~\ref{fig:lcm}.
\vspace{-4pt}
The reason for this phenomenon is that there is no stronger supervisory signal that allows the LCM to produce unblur portrait video, and the LCM ignores the temporal quality of generated videos, even when the loss of the boundary condition is added to the consistency distillation loss.
Therefore, a possible solution is introducing a temporal loss function.
We propose to use a discriminator to provide the adversarial consistency distillation loss to improve the temporal quality.

To be detailed, our discriminator is required to have several characteristics: being able to judge the quality of generation on the temporal, and being able to judge audio-lip consistency. This is key to generating high-quality portrait videos. Some existing works~\cite{mao2024osv} use pre-trained vision models such as the DINO~\cite{caron2021emerging}, which is unsuitable for our setting.
Since we have a trained base model that can generate non-blurry, audio-lip consistency portrait videos given sufficient sampling steps, it is natural to extract the feature map of the base model as part of the discriminator, similar to phased consistency models (PCM)~\cite{wang2024phased}.
The architecture of the discriminator is shown in Fig~\ref{fig:model}.
To be specific, we use the feature maps of temporal attention in downsampling and the middle block in our 3D U-Net, and use 3D convolution to process all the feature maps $h_i$, average the processed feature maps, and obtain the final output of discriminator $D_\varphi(\epsilon_\phi(x_t))$.

Therefore, while training the discriminator, we can use the pairs of ground truth videos and generated videos via latent consistency models using the Eq.~\eqref{eq:advloss}
\begin{align}
\label{eq:advloss}
\mathcal{L}_{\text{dis}}(\phi,x_{\bigtriangleup t},\hat{x}_{\bigtriangleup t};\varphi ) &=\text{ReLU}(1-D_\varphi(\epsilon_\phi(x_{\bigtriangleup t}))) \\ 
& +\text{ReLU}(1+D_\varphi(\epsilon_\phi(\hat{x}_{\bigtriangleup t}))) \nonumber ,
\end{align}
where $\epsilon_\phi$ is the pre-trained teacher diffusion model, $\varphi $ denotes the parameters in discriminator such as the 3D convolution. $x_{\bigtriangleup t}$ and $\hat{x}_{\bigtriangleup t}$ mean the real data and fake or generated data with a small noise level, ${\mathbf x}_{\bigtriangleup t} \sim \mathcal{N}(\alpha_{t}{\mathbf x}_0;\beta^2_{t}\mathbf{I})$, $\hat{\mathbf x}_{\bigtriangleup t} \sim \mathcal{N}(\alpha_{t}\hat{\mathbf x}_0;\beta^2_{t}\mathbf{I})$, respectively.
In our paper, we set the $\bigtriangleup t \in [0,5]$, uniform sampling, which has a better performance than the training while keeping $\bigtriangleup t=0$.

While training latent consistency models, we can use the discriminator to provide the adversarial loss to make the generated samples more realistic,
\begin{align}
\label{eq:advloss2}
\mathcal{L}_{\text{adv}}(\phi,\hat{x}_{\bigtriangleup t};\varphi ) = \text{ReLU}(1-D_\varphi(\epsilon_\phi(\hat{x}_{\bigtriangleup t}))) ,
\end{align}
where $\hat{x}_{\bigtriangleup t}$ is the data prediction of latent consistency models adding noise $\hat{x}_{\bigtriangleup t} = \alpha_{\bigtriangleup t}f_\theta (x_t,t) + \beta_{\bigtriangleup t} \epsilon $.
We define the LCM trained with consistency loss and adversarial loss as the adversarial latent consistency model (Adv-LCM).

\textbf{Progressive forward process for discriminator.}
Inspired by the training trick in the generative adversarial networks~\cite{goodfellow2020generative,gulrajani2017improved}, the speed of convergence of the generator (Adv-LCM in this paper) and the discriminator in the early stages of training can have a large impact on the results. At the beginning of training, Adv-LCM is less effective for one-step inference under large noise levels, so at this time it is less difficult for the discriminator to judge the real and generated samples, leading to the discriminator converging quickly and failing to provide stronger supervision to OSA-LCM. 
Therefore, we propose to use a progressive forward process for discriminator training, which is summarized as,
\begin{align}
t_{\text{dis}} = \text{uniform}(0,\min(\lceil\frac{i}{10}\rceil,1000)),
\end{align}

\begin{figure}[h!]
    \centering
    \begin{subfigure}[b]{0.15\textwidth}
        \centering
        \includegraphics[width=\textwidth]{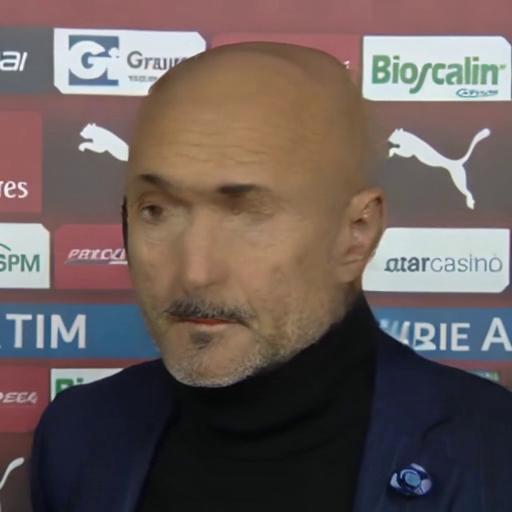}
        \caption{Adv-LCM-1}
    \end{subfigure}
    \begin{subfigure}[b]{0.15\textwidth}
        \centering
        \includegraphics[width=\textwidth]{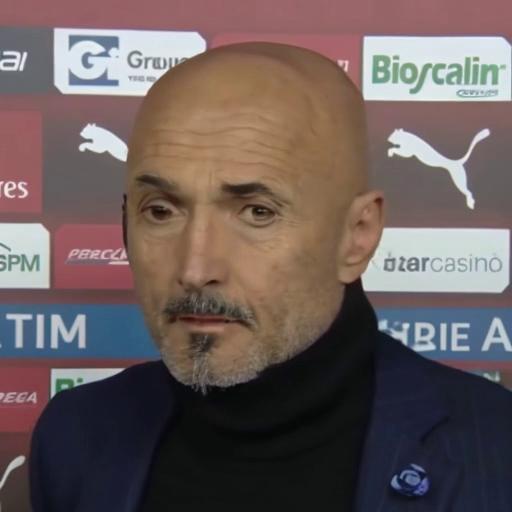}
        \caption{AdvLCM-2}
    \end{subfigure}
    \begin{subfigure}[b]{0.15\textwidth}
        \centering
        \includegraphics[width=\textwidth]{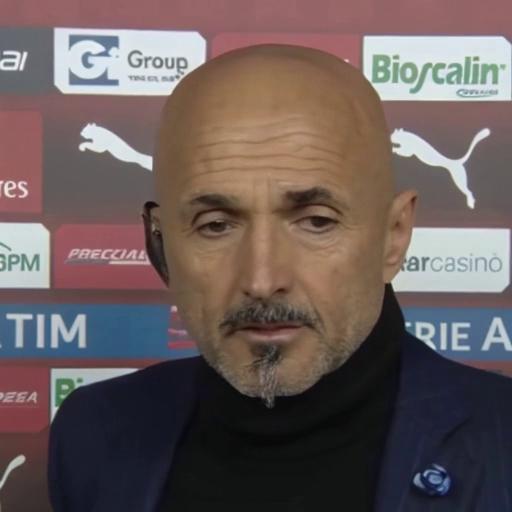}
        \caption{OSA-LCM-1}
    \end{subfigure}
    \caption{\textbf{Different sampling steps for Adv-LCM and OSA-LCM.} Adv-LCM will also generate blurry videos.}
    \label{fig:osalcm}
\end{figure}
\vspace{-.1in}

where $i$ denotes the discriminator training iterations, $\text{uniform}(\cdot)$ is the uniform distribution. While in the iterations training the latent consistency models, the timesteps $t_{\text{lcm}}$ is sampled from $\text{uniform}(0,1000)$ as usual.

\textbf{Motion loss for consistency models.}
In addition, we found that even after adding discriminator adversarial loss, in some scenarios, a slightly blurry situation occurs when the head moves faster, which is because there is no constraint to limit the head's trajectory thus discriminator will guide the OSA-LCM to move in the right direction and therefore reduce the guidance on the motion frames, especially the fast motion frames.
Therefore, we propose an easy motion loss to guide the OSA-LCM about the motion direction to allow the discriminator to maximize its role in guiding the quality of motion frames generated,
\vspace{-5pt}
\begin{align}
\label{motionloss}
\text{diff}(x_0) &= \text{mean} (x_0^{c, f+1, h, w} - x_0^{c, f, h, w}) \\
M(x_0,\hat{x}_0) &= \text{MSE}(\text{diff}(x_0),\text{diff}(\hat{x}_0)) ,
\end{align}
where $x_0$ and $\hat{x}_0$ denote the ground truth video or the generated videos which are all in latent space with shape $(c,F,h,w)$ and $F$ denotes the number of frames.

\textbf{Final objective functions.}
The Stage I training of OSA-LCM consists of two parts, one is to train the latent consistency models and another is to train the discrimination. Referring to the training scheme of the GAN, we optimize the Adv-LCM in one iteration and then optimize the discriminator in the next iteration until convergence.
To optimize the discriminator, we use the discriminative loss in Eq.~\eqref{eq:advloss}.
We use the combined loss objective function for Adv-LCM, 
\vspace{-4pt}
\begin{align}
\mathcal{L}(\theta) & = d(\tilde{\mathbf x}_0, \hat{\mathbf x}_0) + \gamma d(\hat{\mathbf x}_0, x_0) + \\ & \lambda \mathcal{L}_{\text{adv}}(\phi,\hat{x}_{\bigtriangleup t};\varphi)+mM(\hat{\mathbf x}_0,x_0)  , \nonumber
\end{align}
where $d(\cdot)$ denotes the distance metrics taken as the Huber loss, $M(\cdot)$ denotes the motion loss function in Eq.~\eqref{motionloss}, $\tilde{\mathbf x}_{\bigtriangleup t} \sim \mathcal{N}(\alpha_{t}\tilde{\mathbf x}_0;\beta^2_{t}\mathbf{I})$, $\gamma=0.05$, $\lambda=1.0$, $m=0.07$.

Meanwhile, we show the total algorithm of stage I training of OSA-LCM in the following, where $a$ means the audio embedding, $r$ denotes the reference images which are randomly sampled from the whole video and encoded with the VAE, $p$ denotes the past motion frames, which are the n consecutive frames before the start of video $x$. We set the number of past frames equal to 4 in our paper, referred to the EMO~\cite{tian2024emo}. It is noted that when estimating the previous states, Eq.~\eqref{lcm_solver}, in the Adv-LCM, we also use the classifier-free guidance except for the past motion frames, which is beneficial for reducing the error accumulating. 
A schematic of the training loss for stage 1 is shown in Fig.~\ref{fig:onecol}.
Since the animtediff~\cite{guo2023animatediff} framework, we can only train the temporal module to avoid using the ema model as the target model, which can save memory and have better results.

\vspace{-5pt}
\begin{algorithm}[H]
    \begin{minipage}{\linewidth}
    \begin{footnotesize}
        \caption{Stage I training of OSA-LCM}\label{alg:s1}
        \begin{algorithmic}
            \STATE \textbf{Input:} dataset $\mathcal{D}$, teacher model $\epsilon_\theta$, initial OSA-LCM parameter $\theta$, discriminator $D_\varphi$, learning rate $\eta_{\text{lcm}}$ and $\eta_{\text{dis}}$, ODE solver $\Psi(\cdot,\cdot,\cdot, \cdot,\cdot,\cdot)$, distance metric $d(\cdot,\cdot)$, motion metric $M(\cdot,\cdot)$, noise schedule $\alpha_t,\sigma_t$, guidance scale $w=1.2$, number of ODE step $k$, two discrete points separated by $s=1000//k$, loss weight $\gamma$,$\lambda$,$m$. 
            \STATE Training data : $\mathcal{D}_{\mathbf x}=\{(\mathbf x,a,r,p)\}$
            \REPEAT
            \STATE Sample $(x_0,a,r,p) \sim \mathcal{D}_x$, $t\sim \mathcal{U}[0,T]$
            \STATE Sample ${\mathbf x}_{t}\sim \mathcal{N}(\alpha_{t}x_0;\beta^2_{t}\mathbf{I})$
            \STATE Solve
            \STATE $\begin{aligned}{\mathbf x}^{\boldsymbol \phi}_{t-s}\leftarrow (1+\omega)\Psi({\mathbf x}_{t},t,t-s,a,r,p)-\omega\Psi({\mathbf x}_{t},t,t-s,\varnothing,\varnothing,p)\end{aligned}$
            \STATE $\hat{\mathbf x}_0 = \boldsymbol f_{\boldsymbol \theta}(\mathbf x_{t}, t, a,r,p)$ and $\tilde{\mathbf x}_0 = \boldsymbol f_{\boldsymbol{\theta}}(\mathbf x_{t-s}^{\boldsymbol \phi}, t-s, a,r,p)$
            \STATE Sample $\bigtriangleup t \sim \mathcal{U}[0,5]$
            \STATE Sample $\tilde{\mathbf x}_{\bigtriangleup t} \sim \mathcal{N}(\alpha_{t}\tilde{\mathbf x}_0;\beta^2_{t}\mathbf{I})$, $\hat{\mathbf x}_{\bigtriangleup t} \sim \mathcal{N}(\alpha_{t}\hat{\mathbf x}_0;\beta^2_{t}\mathbf{I})$
             \STATE $\begin{aligned}\mathcal{L}(\boldsymbol \theta, \boldsymbol \theta) = & d(\tilde{\mathbf x}_0, \hat{\mathbf x}_0) + \\ &\gamma d(\hat{\mathbf x}_0, x_0) + \lambda \text{ReLU}(1-D_\varphi(\hat{\mathbf x}_0))+mM(\hat{\mathbf x}_0,x_0) \end{aligned}$
          \STATE $\theta\leftarrow\theta-\eta\nabla_\theta\mathcal{L}(\theta)$
            \STATE 
            \STATE Sample $(x_0,a,r,p) \sim \mathcal{D}_x$, $t_{\text{dis}}\sim \mathcal{U}[0,\min(\lceil\frac{i}{10}\rceil,1000)]$
            \STATE Sample ${\mathbf x}_{t_{\text{dis}}}\sim \mathcal{N}(\alpha_{t_{\text{dis}}}x_0;\beta^2_{t_{\text{dis}}}\mathbf{I})$
            \STATE Solve $\hat{\mathbf x}_0 = \boldsymbol f_{\boldsymbol \theta}(\mathbf x_{t_{\text{dis}}}, t_{\text{dis}}, a,r,p)$ 
             \STATE $\begin{aligned}\mathcal{L}_{\text{dis}}(\varphi) = \text{ReLU}(1+x_0)+\text{ReLU}(1-\hat{\mathbf x}_0) \end{aligned}$
             \STATE $\varphi\leftarrow\varphi-\eta\nabla_\varphi\mathcal{L}(\varphi)$
            \UNTIL convergence
        \end{algorithmic} 
        \end{footnotesize}
    \end{minipage}
\end{algorithm}

\subsection{Stage 2: Editing fine-tuned training}

After training the stage 1 of OSA-LCM, we obtained an Adv-LCM which can generate high-quality portrait videos with two sampling steps. 
However, when only one step of sampling is used for Adv-LCM, the quality of the generated portrait videos drops drastically, and the generated videos become blurry, as shown in Fig.~\ref{fig:osalcm}.

The most essential reason for this phenomenon is the presence of the gap between training and inference. At the time of training, the task of Adv-LCM is to predict the data given the noisy data, that is, the $\hat{x}_0 = f_\theta (x_t,t)$. However, while inference in only one step, that $\hat{x}_0 = f_\theta (\epsilon,t)$.
The gap lies in the temporal information between the $\epsilon$ and $x_t$. While training, even noising the data with the maximum noise level $x_T$, there is still a correlation among frames. 

\begin{figure*}
  \centering
    \centerline{\includegraphics[width=0.94\linewidth]{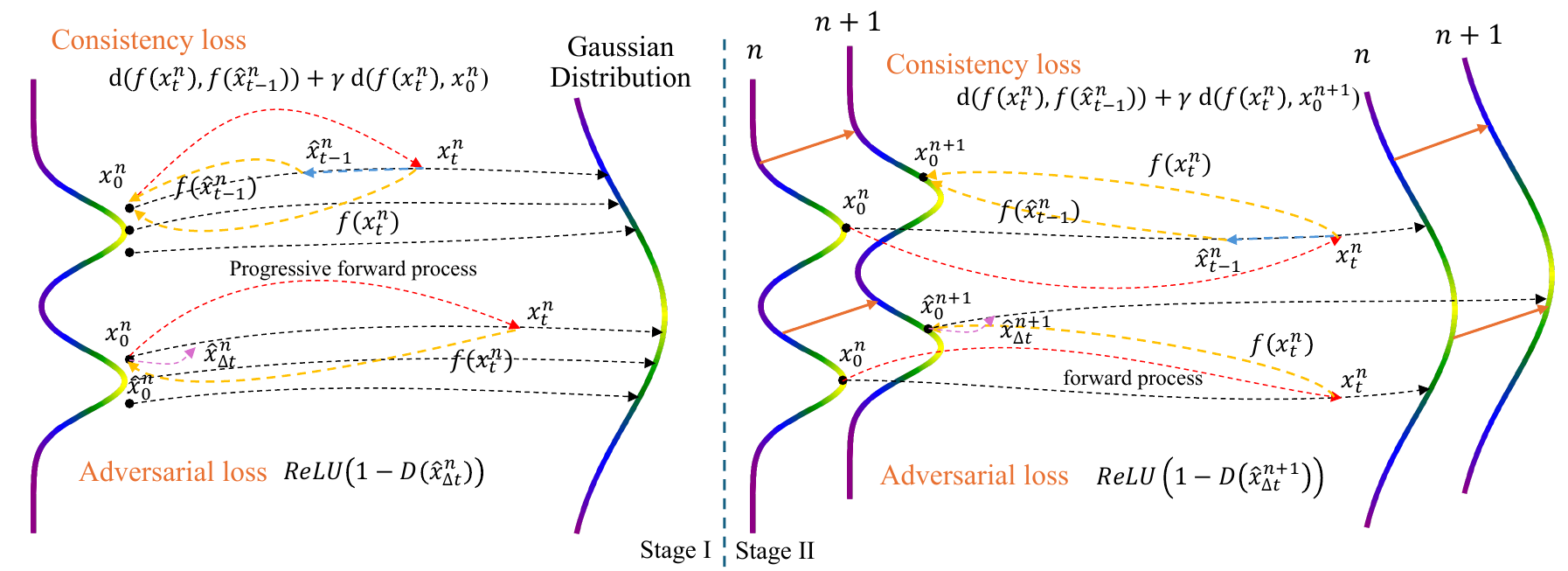}}

   \caption{\textbf{Two stages training of our OSA-LCM.} In the second stage, we propose a novel training scheme that uses the noising data of past frames to predict future frames. Both two stages use adversarial loss and consistency loss.}
   \label{fig:onecol}
\vspace{-.2in}
\end{figure*}

Therefore, during inference, the Adv-LCM's task is to recover the corresponding temporal correlation. 
Since there is no relevant paradigm for the Adv-LCM to change the original input temporal correlation during model training, instead, the current training framework makes the Adv-LCM more reliant on the input temporal correlation. Therefore, in the case where there is only one sampling step, Adv-LCM will be prone to the blurry problem, which is essentially that it does not have a better mapping from the uncorrelated temporal relation in Gaussian distribution.

Motivated by the motion consistency models~\cite{zhai2024motion}, it is possible to reduce the gap by training the LCM with the input of Gaussian noise. 
However, such methods will also bring a challenge if the input of the LCM is only the Gaussian noise, such training is very destabilizing since this is the same as training a GAN.
While balancing the training stability and the performance, an alternative is to sample the Gaussian or noisy data to predict data with some probability.
However, we found that such a training strategy cannot even outperform the default setting, as shown in Tab.~\ref{tab:comparison}.

Therefore, we put forward a novel training strategy that transforms the prediction task into the editing task, called the editing fine-tuned method (EFT). 
To be detailed, we will continuously fine-tune the Adv-LCM to transform from a generation task into an editing task. 
We define the latent of the frames of the n-th fragment of a video to be $x_0^{(n)}$ and an editing model is defined as $x_0^{(n+1)} = E_\theta(x_0^{(n)})$, which is also like one kind of autoregressive model that predicting the next frame or fragment given the frame or fragment of the present. Motivated from the SDEdit~\cite{meng2021sdedit}, to achieve the editing, we can noise the data $x_0^{(n)}$ first and then denoise it to get the next frame or fragment, $x_0^{(n+1)}=H_\theta(x_t^{(n)})$. 

Therefore, after training the Adv-LCM which has an excellent ability to generate the portrait videos $\hat{x}_0=f_\theta(x_t,t)$, we can fine-tune such Adv-LCM to become an editing model, which means that optimize $f_\theta(x_t^{(n)},t) \approx H_\theta(x_t^{(n)},t) $, called editing fine-tuned method (EFT). 
EFT has the following advantages, first is that EFT can give Adv-LCM the ability to change the temporal distribution based on the input of the conditions.
For instance, when there is a difference between the temporal information of the previous video fragment and the next fragment, because of the change in the learning task, Adv-LCM will be more inclined to change the temporal distribution based on the audio condition or other information, which is beneficial for the temporal gap between training and inference.
Another advantage is that we can sample the long videos in the autoregressive format. Once the past videos have been generated, we can extract the generated fragments and noise them to obtain the start of sampling for the next video fragment, called rolling sampling (RS).

\vspace{-4pt}
\begin{algorithm}[H]
    \begin{minipage}{\linewidth}
    \begin{footnotesize}
        \caption{Stage II training of OSA-LCM: EFT}\label{alg:s2}
        \begin{algorithmic}
            \STATE \textbf{Input:} dataset $\mathcal{D}$, teacher model $\epsilon_\theta$, trained OSA-LCM parameter $\theta$, trained discriminator $D_\varphi$, learning rate $\eta_{\text{lcm}}$ and $\eta_{\text{dis}}$, ODE solver $\Psi(...)$, distance metric $d(\cdot,\cdot)$, noise schedule $\alpha_t,\sigma_t$, guidance scale $w$, two discrete points separated by $s=1000//k$, loss weight $\gamma$,$\lambda$. 
            \STATE Training data : $\mathcal{D}_{\mathbf x}=\{(\mathbf x,a,r,p)^{(n)},(\mathbf x,a,r,p)^{(n+1)}\}$
            \REPEAT
            \STATE Sample $(\mathbf x,a,r,p)^{(n)},(\mathbf x,a,r,p)^{(n+1)} \sim \mathcal{D}_x$, $t\sim \mathcal{U}[0.8T,T]$
            \STATE Sample ${\mathbf x}_{t}^{(n)}\sim \mathcal{N}(\alpha_{t}x_0^{(n)};\beta^2_{t}\mathbf{I})$
            \STATE Solve
            \STATE $\begin{aligned}{\mathbf x}^{(n)}_{t-s;\boldsymbol \phi}\leftarrow &(1+\omega)\Psi({\mathbf x}_{t}^{(n)},t,t-s,a^{(n)},r^{(n)},p^{(n)})-\\&\omega\Psi({\mathbf x}_{t}^{(n)},t,t-s,\varnothing,\varnothing,p^{(n)})\end{aligned}$
            \STATE $\hat{\mathbf x}_0^{(n+1)} = \boldsymbol f_{\boldsymbol \theta}(\mathbf x_{t}^{(n)}, t^{(n)}, a^{(n)},r^{(n)},p^{(n)})$ 
            \STATE $\tilde{\mathbf x}_0^{(n+1)} = \boldsymbol f_{\boldsymbol{\theta}}(\mathbf x^{(n)}_{t-s;\boldsymbol \phi}, t-s, a^{(n)},r^{(n)},p^{(n)})$
            \STATE Sample $\bigtriangleup t \sim \mathcal{U}[0,5]$
            \STATE Sample $\hat{\mathbf x}^{(n+1)}_{\bigtriangleup t} \sim \mathcal{N}(\alpha_{t}\hat{\mathbf x}^{(n+1)}_0;\beta^2_{t}\mathbf{I})$
             \STATE $\begin{aligned}\mathcal{L}(\boldsymbol \theta, \boldsymbol \theta) = & d(\tilde{\mathbf x}^{(n+1)}_0, \hat{\mathbf x}^{(n+1)}_0) + \\ &\gamma d(\hat{\mathbf x}^{(n+1)}_0, x_0^{(n+1)}) + \lambda \text{ReLU}(1-D_\varphi(\hat{\mathbf x}_0^{(n+1)})) \end{aligned}$
          \STATE $\theta\leftarrow\theta-\eta\nabla_\theta\mathcal{L}(\theta)$
            \STATE 
            \STATE $(x_0,a,r,p)^{(n)},(x_0,a,r,p)^{(n+1)} \sim \mathcal{D}_x$, $t_{\text{dis}}\sim \mathcal{U}[0.8T,T]$
            \STATE Sample ${\mathbf x}_{t_{\text{dis}}}^{(n)} \sim \mathcal{N}(\alpha_{t_{\text{dis}}}x_0^{(n)};\beta^2_{t_{\text{dis}}}\mathbf{I})$
            \STATE Solve $\hat{\mathbf x}_0^{(n+1)} = \boldsymbol f_{\boldsymbol \theta}(\mathbf x^{(n)}_{t_{\text{dis}}}, t_{\text{dis}}, a^{(n)},r^{(n)},p^{(n)})$ 
             \STATE $\begin{aligned}\mathcal{L}_{\text{dis}}(\varphi) = \text{ReLU}(1+x_0^{(n+1)})+\text{ReLU}(1-\hat{\mathbf x}^{(n+1)}_0) \end{aligned}$
             \STATE $\varphi\leftarrow\varphi-\eta\nabla_\varphi\mathcal{L}(\varphi)$
            \UNTIL convergence
        \end{algorithmic} 
        \end{footnotesize}
    \end{minipage}
\end{algorithm}

\textbf{Objective functions for EFT.}
We will inherit the Adv-LCM $F_\theta(\cdot)$ trained in the previous stage as the initialization for this stage and re-parametric the Adv-LCM as,
\begin{align}
\label{cm_eq2} 
f_\theta(x_t^{(n)},t) = c_{\text{skip}}(t) x_t^{(n)} + c_{\text{out}}(t) F_\theta(x_t^{(n)},t),
\end{align}
where n denotes the n-th fragment in a video.
The loss is,
\begin{align}
\scalebox{0.68}{%
$\mathcal{L}(\theta)_{\text{EFT}} = d(\tilde{\mathbf x}_0^{(n+1)}, \hat{\mathbf x}_0^{(n+1)}) + \gamma d(\hat{\mathbf x}_0^{(n+1)}, x_0^{(n+1)}) +  \lambda \mathcal{L}_{\text{adv}}(\phi,\hat{x}_{\bigtriangleup t}^{(n+1)};\varphi),$
}
\end{align}
where $x_0^{(n+1)}$ is the ground truth latent of (n+1)-th video fragment, $\hat{\mathbf x}_0^{(n+1)} = f_\theta(x_t^{(n)},t)$ denotes the output of one-step editing model, $\tilde{\mathbf x}_0^{(n+1)} = f_\theta(x_{t-\bigtriangleup  t;\phi}^{(n)},t-\bigtriangleup  t)$
where $x_{t-\bigtriangleup  t;\phi}^{(n)}$ is the estimation of the previous state by pre-defined ODE solver $\Phi$ in the empirical PF-ODE trajectory,
\begin{align}
\label{lcm_solver2}
x_{t-\bigtriangleup  t;\phi}^{(n)} = x_t^{(n)} - \bigtriangleup  t \cdot  \Phi(x_t^{(n)},t;\phi ),
\end{align}
and $\hat{x}_{\bigtriangleup t}^{(n+1)} \sim \mathcal{N}(\alpha_{\bigtriangleup t}\hat{x}_{0}^{(n+1)},\beta^2_{\bigtriangleup t}I)$.
However, due to the parameterized form of Eq.~\eqref{cm_eq2} and $c_{\text{skip}}(0)=1$, $c_{\text{out}}(0)=0$, it means that $f_\theta(x_0^{(n)},0)=x_t^{(n)} \ne x_t^{(n+1)} $.
Therefore, in the EFT, we will sample timestep $t \sim \mathcal{U}[0.8T,T]$ instead of from $\mathcal{U}[0,T]$, which is also more in line with the SDEdit idea.
We also show the algorithm of EFT in the Algo.~\ref{alg:s2}, while a specific EFT process schematic is shown in Fig.~\ref{fig:onecol}.

Meanwhile, in both stages of training, we will make a constraint on the start margin of the video, i.e., we will not sample the first 7 frames of the video in our paper. For the stage of EFT, we will let $x_0^n$ and $x_0^{n+1}$ have the same number of frames so that $x_0^n$ can be used directly as the input to OSA-LCM after noising, but due to the sampling of the start index, there may not be enough frames for $x_0^n$. 
Therefore, when the number of frames in $x_0^n$ is less than $x_0^{n+1}$, we will repeat the latent of the reference image n times and place it in front of $x_0^n$.

\section{Experiments}

\subsection{Experiments setups}
\label{sec:expsetup}
All the experiments about the OSA-LCM, Adv-LCM, and LCM were conducted on a GPU server equipped with 8 NVIDIA H100 GPUs. 
Similar to the training of EMO~\cite{tian2024emo}, we train the base using the same two-stage training with the HDTF and VFHQ dataset by adding some collected anime data. After training the base model, we will use the proposed two-stage training method to further distillate from the base model to obtain the OSA-LCM.

The training process of OSA-LCM was executed in two stages. In the first stage, we train the Adv-LCM based on the latent consistency models with a discriminator, which comprised 30,000 iterations with a batch size of 1 per GPU, utilizing videos with a resolution of 512 × 512 pixels. In the second stage, we use the proposed editing fine-tuned method (EFT) to further train the Adv-LCM to obtain the OSA-LCM, which comprised 15,000 iterations with a batch size of 1, targeting a video resolution of 512 × 512 pixels. 
We use 5000 clips from the HDTF dataset filter with the audio-lipped consistency and 300 collected anime clips to train the latent consistency models in two stages of training of the OSA-LCM. 
Meanwhile, we trained the Adv-LCM or LCM with a decay learning rate which equals 3e-6 in the half of training and 1.5e-6 in the rest, trained the OSA-LCM with a constant learning rate equals 1.5e-6 while the learning rate of discriminator sets as 1e-5.

\subsection{Quantitative Comparisons}\label{secquan}

In this section, we will evaluate the quality of generated portrait videos under different methods and different sampling steps by widely used metrics, the FID~\cite{heusel2017gans} and the FVD~\cite{unterthiner2019fvd}. 
We don't report other metrics such as the Sync~\cite{chung2017out} since the audio-lip synchronization is excellent among all the methods and isn't challenging for distillation tasks. 
The main challenge for acceleration is the quality of generated portrait videos, including the quality of the generated video in or among frames.

We report the results in Tab.~\ref{quan}, for the baseline of LCM, we carry out several settings. 
The first one is the vanilla LCM, which is similar to VideoLCM~\cite{wang2023videolcm}, denotes as LCM and we use the motion consistency models (MCM)~\cite{zhai2024motion}. Furthermore, we report the results of Adv-LCM which is trained the LCM with the consistency loss and our proposed adversarial loss.
We carry out several second-stage training settings for Adv-LCM to further improve its performance on limited sampling steps. Adv-LCM (+MCM) means that after training the Adv-LCM, we fine-tuned it using the training of the MCM but sampling timestep from $\mathcal{U}[0.8T,T]$, while Adv-LCM (+FT) means fine-tuned Adv-LCM with timestep from $\mathcal{U}[0.8T,T]$ without change the objective function.
Adv (+EFT/RS) means continuously fine-tuning the Adv-LCM using the EFT while inference in rolling sampling. OSA-LCM means fine-tuning the Adv-LCM using the EFT and inference from Gaussian. 

The quantitative result is to randomly select 100 clips from HDTF and VFHQ respectively and then randomly select one frame as reference images and the audio of the whole video according to fps=15 to generate videos and calculate FID and FVD with the original 200 videos.

For the Tab.~\ref{tab:comparison}, our base model outperforms the open-source model, which makes the distillation model trained on it more convincing and practically applicable. 
With limited sampling steps, our proposed Adv-LCM can also outperform LCM and MCM with two or four sampling steps while keeping a similar performance to the base model.
Furthermore, OSA-LCM fine-tuned on the Adv-LCM outperforms other training settings in one step and it's basically on par with the base model.

\begin{table}[t]
\centering
\caption{Quantitative results comparison on HDTF/VFHQ. Videos are saved at 15 fps. NFE is the number of function evaluations.}
\label{tab:comparison}
\begin{tabular}{lccc}
\toprule
\textbf{Method} & \textbf{NFE} & \textbf{FID} & \textbf{FVD} \\
\midrule
OSA-LCM & 1 & \textbf{23.54} & \textbf{124.76} \\
Adv-LCM (+EFT/RS)& 1 & 34.18 & 159.35 \\
Adv-LCM (+FT)& 1 & 24.30 & 137.24 \\
Adv-LCM & 1 & 24.85  & 143.36 \\
LCM & 1 & 39.09 & 324.00 \\
Adv-LCM (+MCM) & 1 & 25.99 & 141.80 \\
MCM & 1 & 26.03 & 145.57 \\
\midrule
Adv-LCM & 2 & \textbf{22.98} & \textbf{128.87} \\
LCM & 2 & 27.06 & 174.31 \\
\midrule
Adv-LCM & 4 & \textbf{23.39} & \textbf{133.80} \\
LCM & 4 & 25.00 & 143.02 \\
Echomimic~\cite{chen2024echomimic} & 4 & 33.23 & 223.42 \\
\midrule
\textbf{Ours (base)} & 20 & \textbf{23.40} & \textbf{120.64} \\
Hallo~\cite{xu2024hallo} & 40 & 49.27 & 248.21 \\
Echomimic~\cite{chen2024echomimic} & 20 & 29.22 & 176.15 \\
Aniportrait~\cite{wei2024aniportrait} & 20 & 30.74 & 192.23 \\
\bottomrule
\label{quan}
\end{tabular}
\vspace{-.4in}
\end{table}
\begin{figure*}
  \centering
    \centerline{\includegraphics[width=0.98\linewidth]{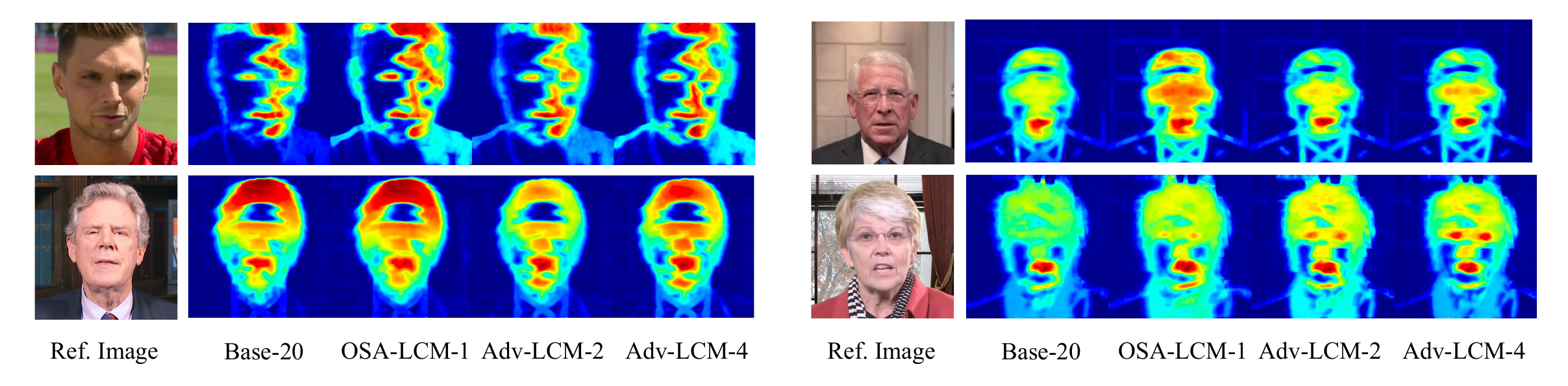}}

   \caption{\textbf{Motion of heatmap for different sampling steps.} It is shown that our Adv-LCM or OSA-LCM can generate a similar motion compared with the base model, and even the motion of OSA-LCM is bigger in some cases. Adv-LCM-4 denotes the heatmap of portrait videos generated by using Adv-LCM with four sampling steps.}
   \label{fig:heatmap}
\vspace{-.2in}
\end{figure*}
\subsection{Qualitative Comparisons}
Besides the quality of generated portrait videos, another essential issue in inference acceleration is whether the motion and expressivity are reduced without sufficient sampling steps. 
In this section, we will use the heatmap to analyze the motion and expressivity for different sampling steps. 
To be detailed, for each pixel in one generated video, we will calculate its optical flow among two frames, accumulate the optical flow of the whole video, and then display the value of accumulated flow with a heatmap, so the redder the heatmap comes out of a certain area means the bigger the motion in that area.

Fig.~\ref{fig:heatmap} shows that the Adv-LCM with four steps generates portrait videos with a similar motion to the base model, but the motion magnitude decreases when the number of inference steps decreases to two, while OSA-LCM further fine-tuned with EFT generates portrait videos with similar or even larger motion than the base model at only one sampling step. Although there is a decrease in the generation quality (Sec.~\ref{secquan}), Fig.~\ref{fig:heatmap} illustrates that EFT can mitigate the gap between training and inference for temporal modeling, thus allowing it to perform better in temporal generation.

\subsection{Ablation Studies}

We carry out the comparison between the different forward processes in training discriminators for the Adv-LCM. Tab.~\ref{tab:comparison1} shows that the Adv-LCM generation is better because the progressive forward makes the discriminator's discrimination more difficult and thus provides better supervision.

Meanwhile, we figured out that adding the motion loss to the Adv-LCM will let the discriminator more focus on the quality of motion instead of the direction of motion, it will also improve the generated performance of the Adv-LCM.

\begin{table}[t]
\centering
\caption{Quantitative results of Adv-LCM between different training settings. NFE is the number of function evaluations. }
\label{tab:comparison1}
\begin{tabular}{lccc}
\toprule
\textbf{Method} & \textbf{NFE} & \textbf{FID} & \textbf{FVD} \\
\midrule
base Adv-LCM & 2 & 25.21 & 152.33 \\
+ progressive forward & 2 & 24.31 & 144.02 \\
+ motion loss & 2 & 22.98 & 128.87 \\
\bottomrule
\end{tabular}
\vspace{-.25in}
\label{table:aba_dis}
\end{table}
\vspace{-.1in}

\section{Conclusion}

In this paper, we introduce the OSA-LCM, which utilizes a two-stage training process to enhance the stability and efficiency of portrait video generation.
In both stages, we employ GAN-like training with novel discriminator design. 
With only one training stage, base Adv-LCM can generate high-quality with two sampling steps. However, it fails to generate natural and unblurry videos with one sampling step and therefore we propose a novel training method called EFT to further fine-tune the Adv-LCM in editing modeling.
Our experiments demonstrate the results of using the OSA-LCM with one step can be similar to using the OSA-LCM with one step, which realizes a 20x times speedup.
By using OSA-LCM in Nvidia-H100, it takes less than 1s to generate one second of high-quality portrait videos.

{
    \small
    \bibliographystyle{ieeenat_fullname}
    \bibliography{main}
}
\clearpage
\setcounter{page}{1}
\maketitlesupplementary

\section{More discussion about the rolling sampling}
\label{sec:rs}
We have discussed one of the advantages of our proposed EFT: It can use rolling sampling, which adds noise to the past n frames, and utilize the noisy data as the start of sampling instead of the Gaussian, which is one kind of auto-regressive. 
However, in the experiment, we found that rolling sampling for the Adv-LCM performed worse than the base Adv-LCM in one sampling step.

We went deep into the reason why the effect of using rolling sampling in Adv-LCM after EFT decreases so much, and then it was found that the main reason is the effect of the first fragment of generated video. 
Since we need to noise the past frames as the initialization for the next video generation, but the first fragment of the video does not have past frames, we need to noise the reference image which repeated n (n is 26) times, although in the training we will likewise consider when the number of past frames is not enough, it will use the repeated reference image, but the probability without any past frames is extremely low, thus leading to this will produce a larger problem, see Fig.~\ref{fig:copypast}, when using the reference image repeated RS, will produce the problem of ghosting.

\begin{figure}[h!]
  \centering
    \centerline{\includegraphics[width=0.98\linewidth]{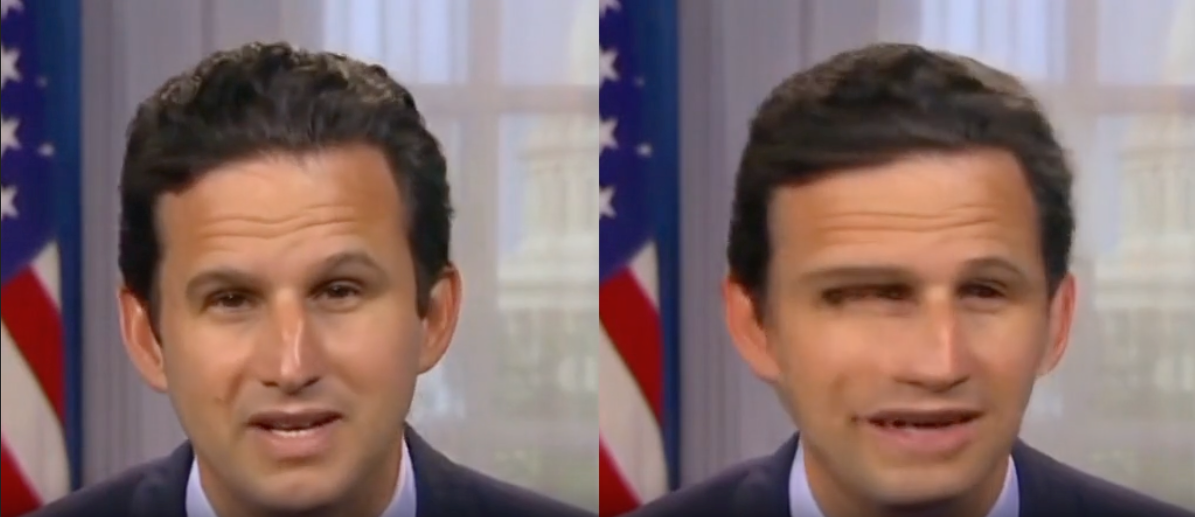}}

   \caption{\textbf{Comparison between the vanilla sampling and rolling sampling.} Left: Vanilla sampling. Right: Rolling sampling. There is a severe ghosting phenomenon, caused by the redundancy of information due to the repeat of the reference image in the first fragment of generated video.}
   \label{fig:copypast}
\end{figure}

Therefore, in order to improve the performance of rolling sampling, we use a combined sampling strategy, which for the first fragment of videos to generate, will use the Gaussian to start with, and for the rest of the videos to generate, it will use the noisy past frames (noise level to control the start point) to start with, called combined rolling sampling.
The Tab.~\ref{table:supp_sample} reports the performance of different combined rolling sampling, it is concluded that the combined rolling sampling can resolve the artifacts brought from the repeat of reference image and improve the performance. 
Meanwhile, with the noise level adding the past frame decreasing, the FID is decreased, which means that the quality of the image dimension is better, which is due to the closer proximity to the reference image with a smaller noise level. Still, the FVD will be worse, which means that the naturalness of the generated video decreases.
Meanwhile, using combined rolling sampling can still bring some advantages, such as better identity consistency, as can be seen in Fig.~\ref{fig:copypast2} some details do not appear out of thin air when using rolling sampling, such as the hair on the right side of Fig~\ref{fig:copypast2}. It is, therefore, a trade-off between identity consistency and video quality or the motion expressivity.

\begin{figure}[h!]
  \centering
    \centerline{\includegraphics[width=0.98\linewidth]{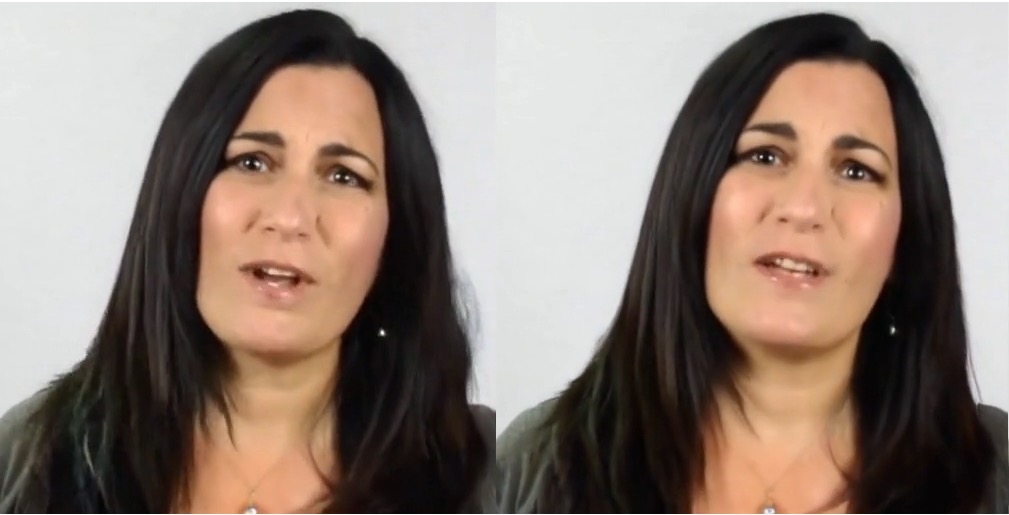}}

   \caption{\textbf{The motion heatmap of different sampling strategies.} With different sampling strategies, the motion of generated portrait videos is similar, which means the principle of different sampling strategies is similar and correct.}
   \label{fig:copypast2}
\end{figure}

Furthermore, we research the motion of generated portrait videos under different sampling steps, from Fig.~\ref{fig:heatmap}, we figure out that using different sampling strategies, the motions of the generated portrait videos are similar, which is in line with our expectations.

\begin{figure}[h!]
  \centering
    \centerline{\includegraphics[width=0.98\linewidth]{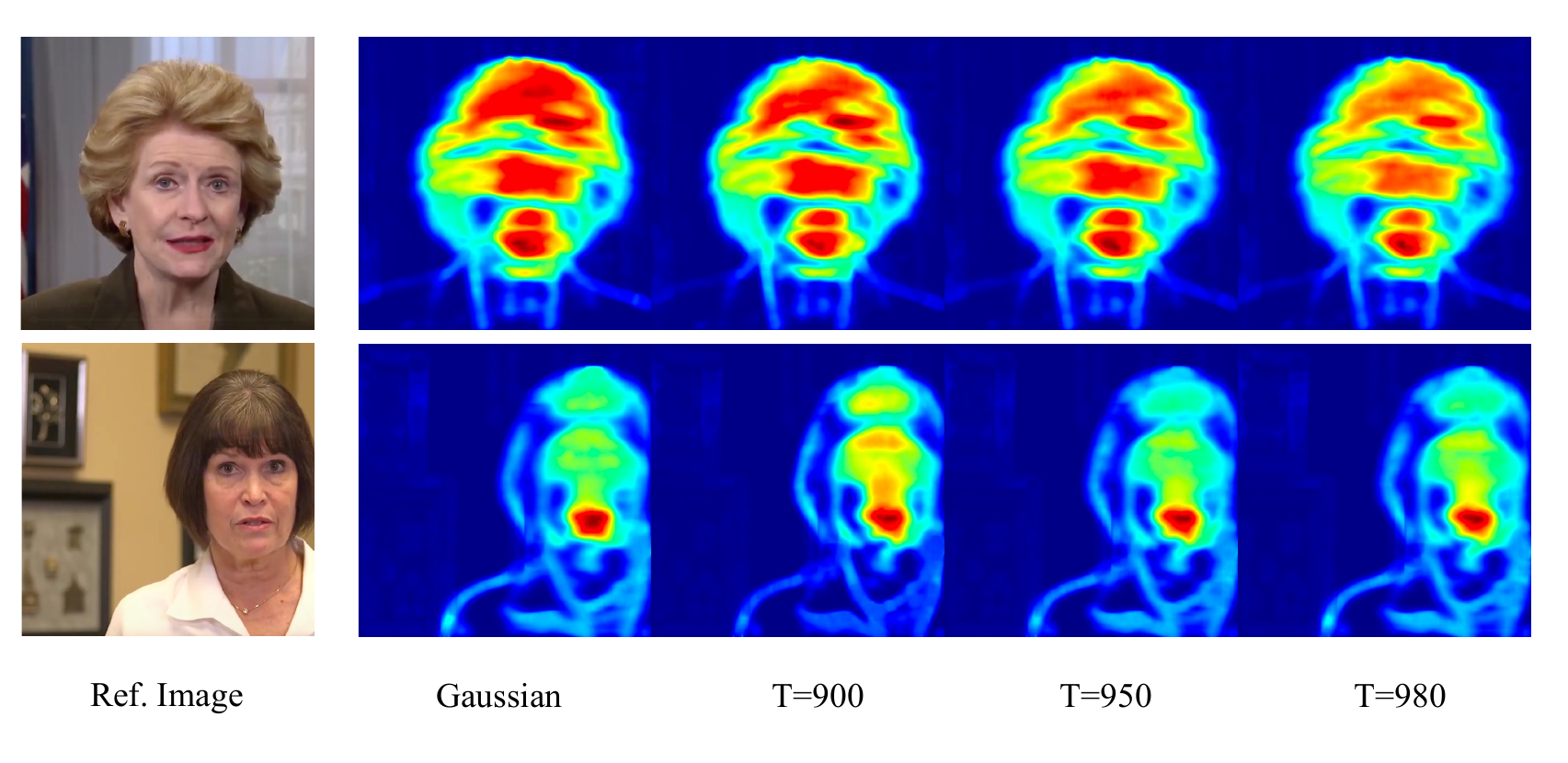}}
   \caption{\textbf{Comparison between the vanilla sampling and rolling sampling.} Left: Vanilla sampling. Right: Rolling sampling. There is a severe ghosting phenomenon, caused by the redundancy of information due to the repeat of the reference image in the first fragment of generated video. Meanwhile, T denotes the noise level of sampling initialization.}
   \label{fig:heatmap}
\end{figure}

\begin{table}[t]
\centering
\caption{Quantitative results of OSA-LCM with different noise levels to start with in sampling. All the settings will sample from Gaussian in the first fragment of generated videos.}
\label{tab:comparison1}
\begin{tabular}{lccc}
\toprule
\textbf{Method} & \textbf{NFE} & \textbf{FID} & \textbf{FVD} \\
\midrule
Sampling from Gaussian & 1 & 23.54 & \textbf{124.76} \\
noise level = 980 & 1 & 22.50 & 125.88 \\
noise level = 950 & 1 & 23.16 & 131.22 \\
noise level = 900 & 1 & 21.86 & 138.12 \\
noise level = 850 & 1 & \textbf{21.36} & 130.55 \\
\bottomrule
\end{tabular}
\vspace{-.25in}
\label{table:supp_sample}
\end{table}

Meanwhile, another interesting issue we haven't discussed is why not use the denoising diffusion bridge models (DDBM)~\cite{zhou2023denoising} to model the second stage training.
This is because, for the SDEdit-based approach, he modeled the edge of the score function $s_\theta = \nabla \log p(x_t,t)$, while for the DDBM, it modeled the conditional score function $s_\theta = \nabla \log p(x_t,t|x_0,x_T)$.
The trained Adv-LCM on the first stage or the base model used is a first-kind model, and therefore, fine-tuned with similar modeling, it is easier to train and has better performance.

\section{More comparison between different sampling steps}

For the sake of experimental completeness, we show in Fig.~\ref{fig:copypast} and Fig.~\ref{fig:copypast2} the generation results of our method under unsynchronized counts, specifically using our base model with the DDIM scheduler~\cite{song2020denoising}, the Adv-LCM with LCM scheduler~\cite{song2023consistency}, OSA-LCM with LCM scheduler (with Gaussian distribution as initialization of sampling by default).

\begin{figure*}[h!]
  \centering
    \centerline{\includegraphics[width=0.98\linewidth]{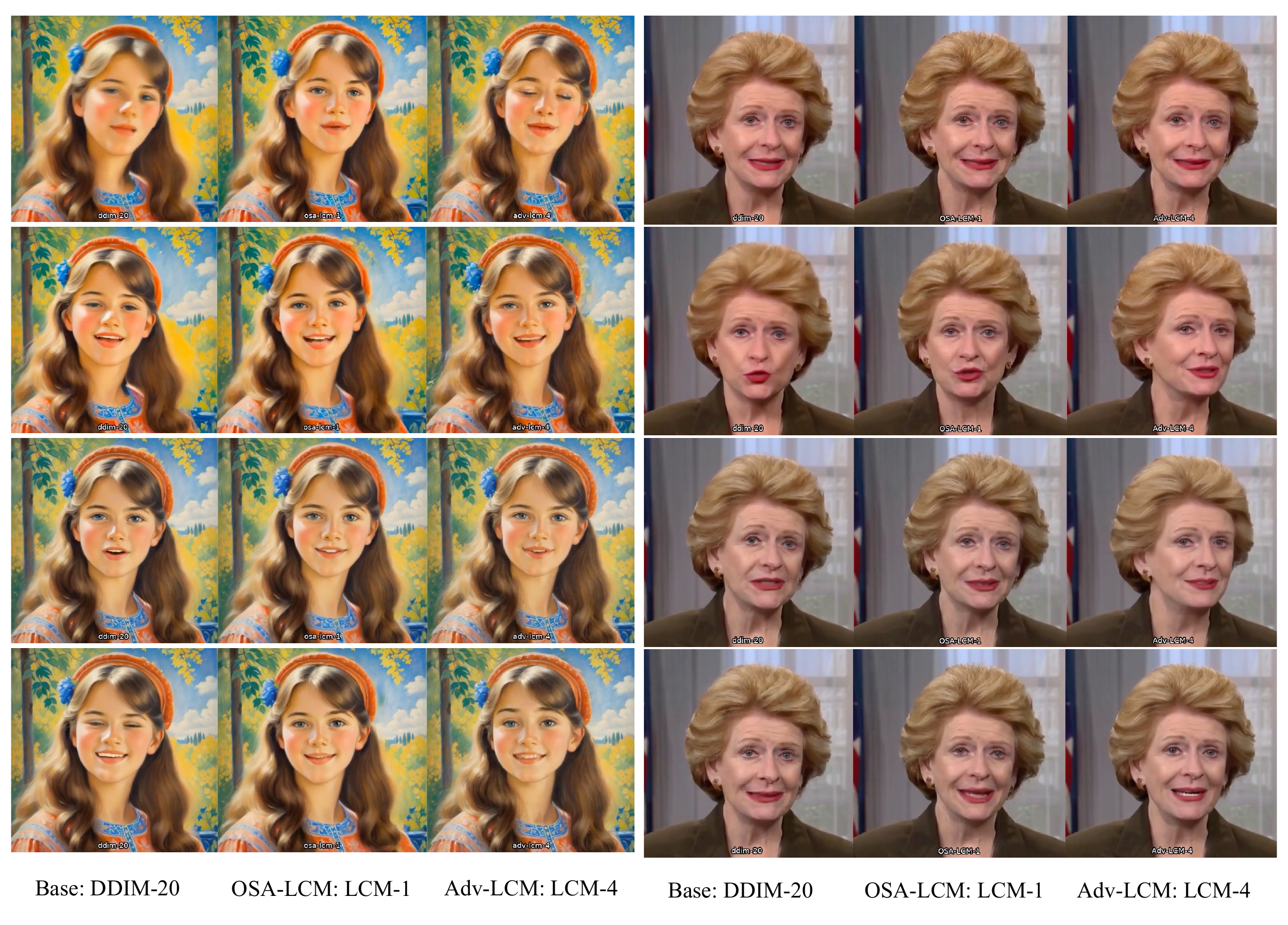}}

   \caption{\textbf{Comparison between the different sampling steps.} The results of the generated portrait video via OSA-LCM with one sampling step show similar quality and motion expressivity with the generated videos via the base model with 20 sampling steps, which means a speed-up of 20x times.}
   \label{fig:copypast}
\end{figure*}

Fig.~\ref{fig:copypast} shows the performance of default number function evaluation (NFE) with different models. 
From Fig.~\ref{fig:copypast}, it can be seen that the result of OSA-LCM using one step is very similar to that of base model reasoning with 20 steps in terms of the quality of the generated video and terms of the motion, which illustrates the usefulness of OSA-LCM.

Meanwhile, Fig.~\ref{fig:copypast2} shows that generated portrait video for all models in sampling only one step with LCM scheduler.
For base model inference only one step generates portrait videos that are completely mushy, for Adv-LCM inference one step generates portrait videos that will have more artifacts, even if for Adv-LCM with two-stage FT (the fine-tuning strategy mentioned in the main text of adding purely Gaussian noise inputs to predict videos) also does not solve the problem of generation quality better.
For OSA-LCM, the fine-tuning strategy of EFT is used, which shows more robustness on one-step inference, indicating that EFT helps to solve the problem of the temporal gap between training and inference during one-step generation.
We put the video in an attached file.

\begin{figure*}[h!]
  \centering
    \centerline{\includegraphics[width=0.98\linewidth]{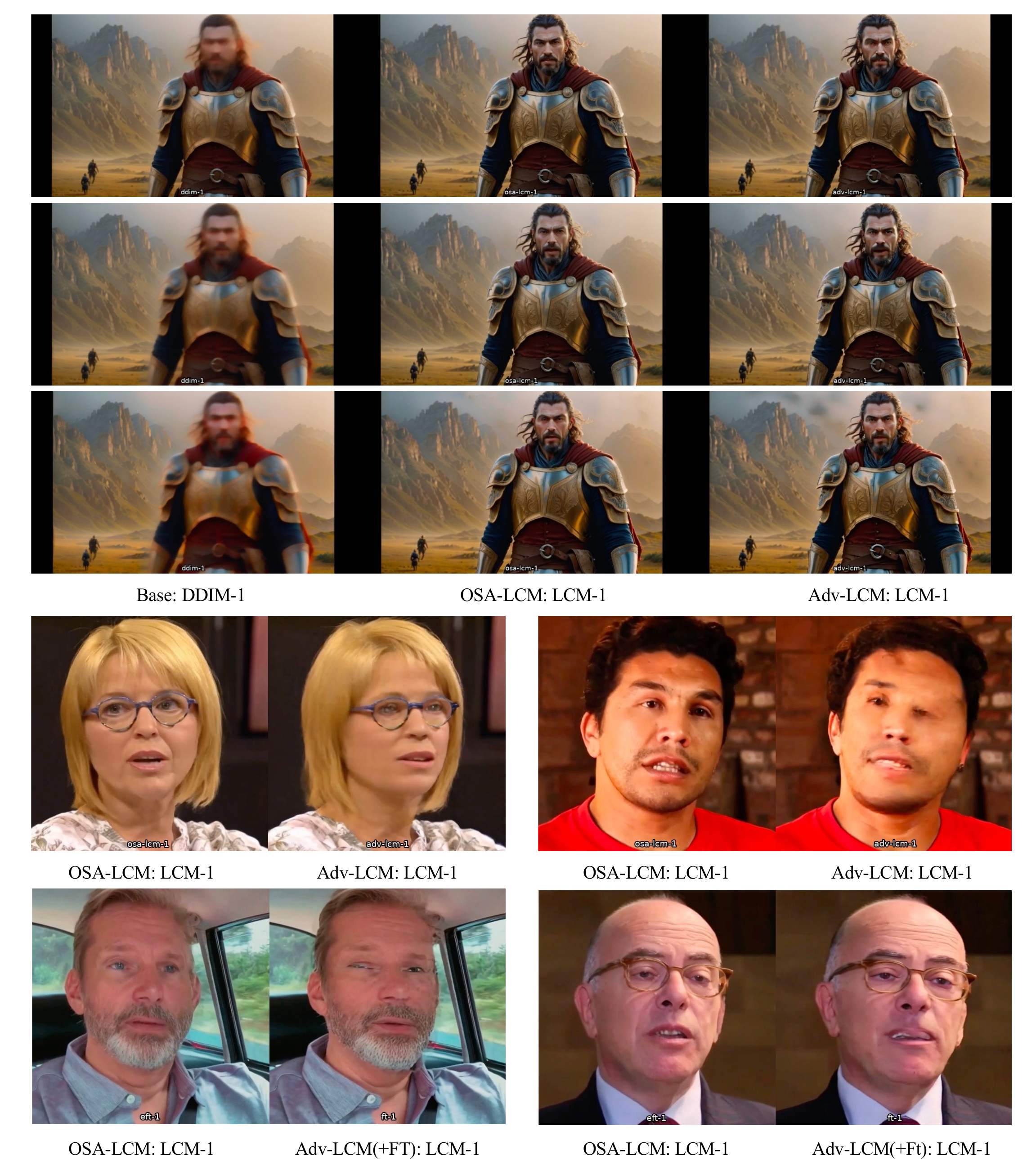}}

   \caption{\textbf{Comparison among different models with only one sampling step.} OSA-LCM can generate high-quality of portrait videos with one sampling step, while other models will generate different types of artifacts with limited sampling steps.}
   \label{fig:copypast2}
\end{figure*}

\section{More comparison between different methods}

Furthermore, for the sake of completeness, we also carry out a comparison of our OSA-LCM with other methods, specifically, we utilized Hallo to reason 40 steps using the DDIM scheduler, and Echomimic's accelerated version (Echo-Acc) to reason 1 or 6 steps using the LCM scheduler (their open-source default setting).
In addition, in this section in order to show the robustness of different approaches to audio types, we will choose to use songs as the audio to drive.

Fig.~\ref{fig:supp} shows the performance of generated portrait videos via different methods with different sampling steps.
The driving of the singing audio is not a problem for any of the different methods, but as can be seen from Fig.~\ref{fig:supp}, OSA-LCM's generation under one step of inference will yield better results than several other methods with more sampling steps, in terms of fewer artifacts as well as a more natural motion.

When these methods reduce the number of sampling steps, the quality of the generated portrait videos drops drastically, even for distillation models like Echo-Acc. When Echo-Acc uses only one sampling step, its generated portrait videos are also completely mushy, indicating that it cannot generate with very small sampling steps.
In addition, other methods are unable to drive on anime data due to the need to detect the face first to determine the position or mask, while our OSA-LCM can also support the driving of anime's reference image.
OSA-LCM shows robust results among different reference images and different audio types with only one sampling step.
We put the video in an attached file.

\begin{figure*}[h!]
  \centering
    \centerline{\includegraphics[width=0.98\linewidth]{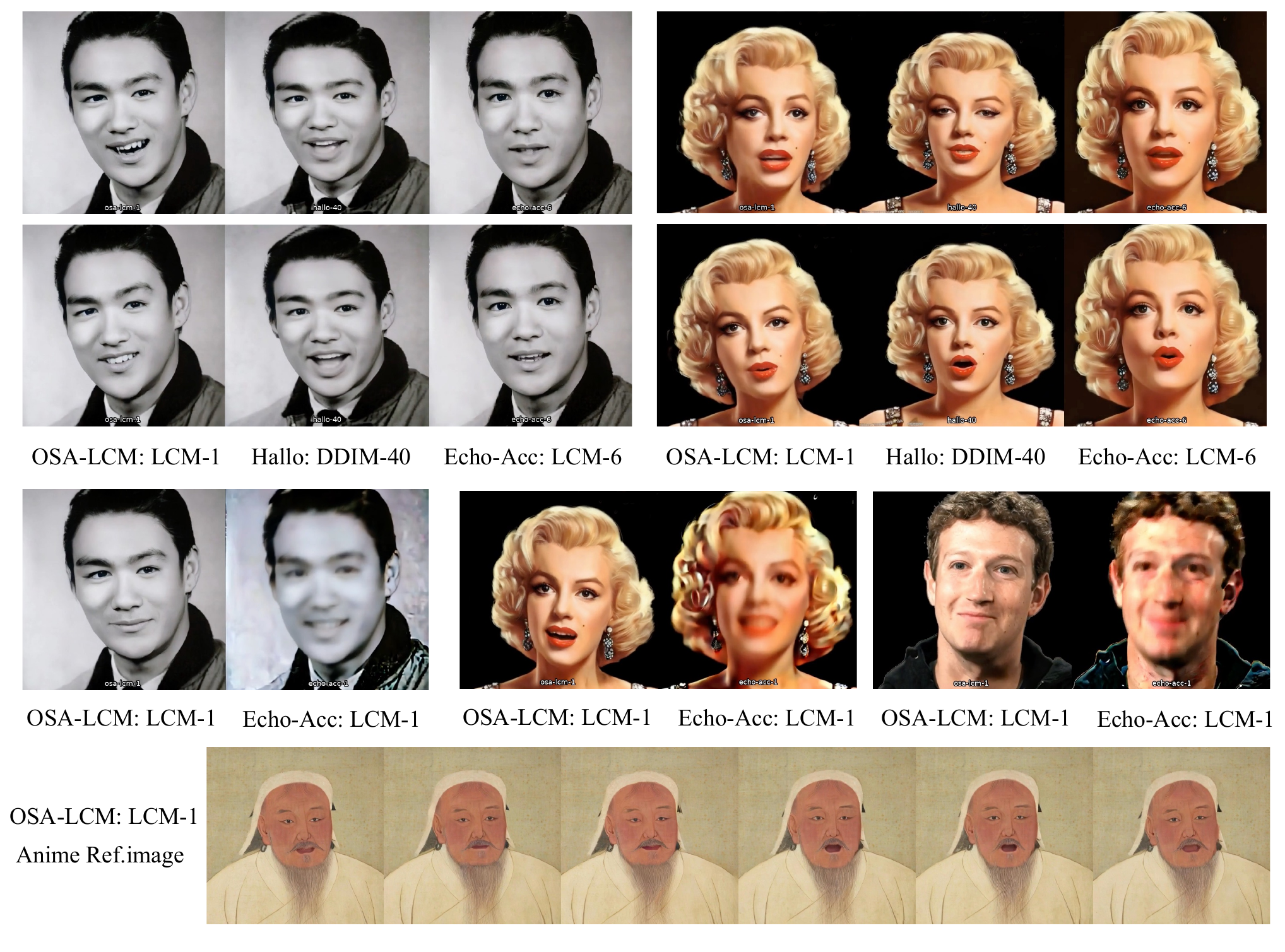}}

   \caption{\textbf{Comparison with different methods of generated portrait videos based on diffusion models.} 
   OSA-LCM can outperform other portrait video diffusion models while using only one step of sampling and shows strong robustness over different audio types and reference images, demonstrating the possibilities in practical use.}
   \label{fig:supp}
\end{figure*}

\begin{figure*}[h!]
  \centering
    \centerline{\includegraphics[width=0.98\linewidth]{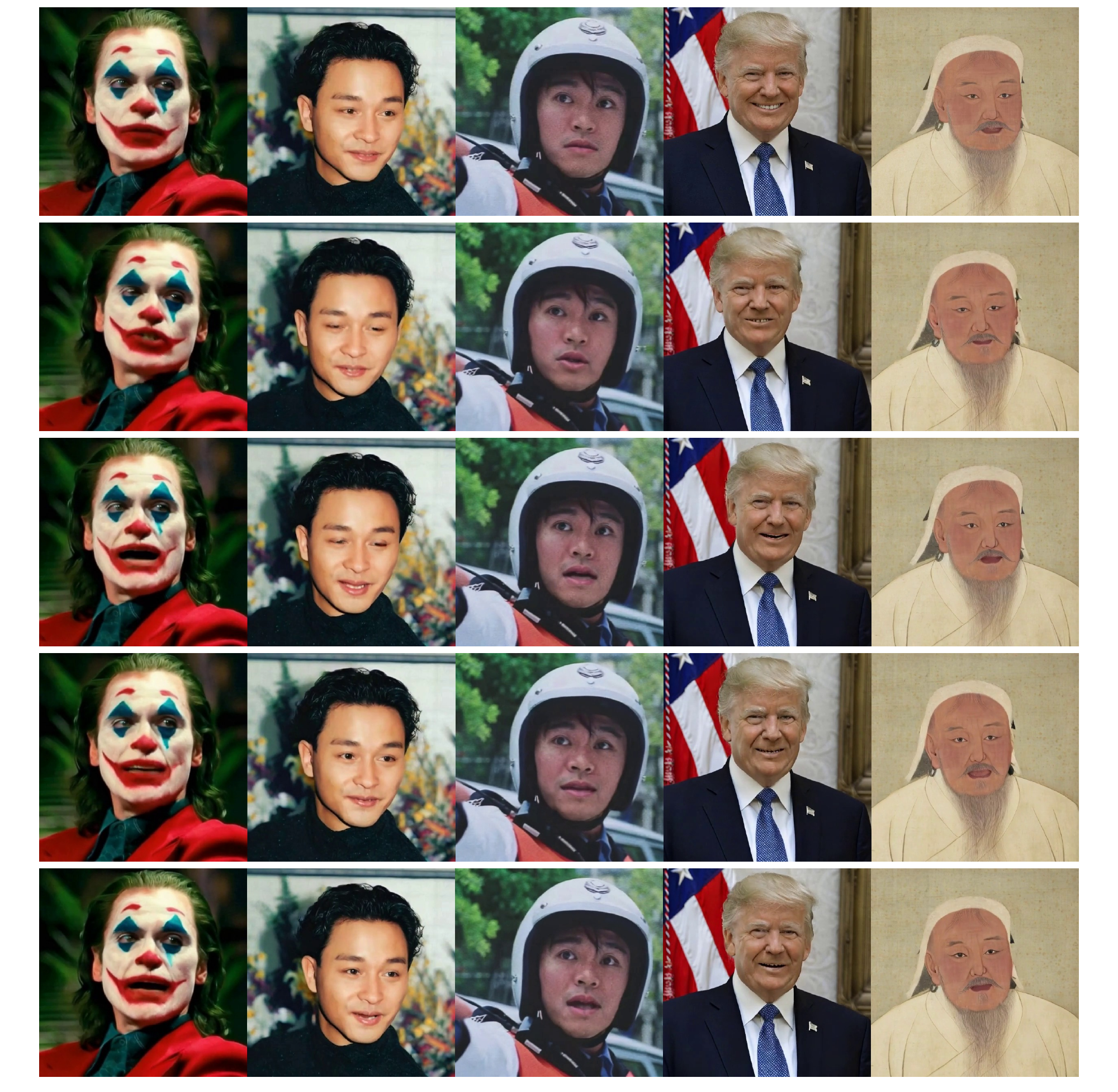}}

   \caption{\textbf{More results of OSA-LCM with one sampling step.} }
   \label{fig:supp}
\end{figure*}

\end{document}